\documentclass{bmvc2k}

\title{Making a Science of Model Search}

\addauthor{J. Bergstra}{bergstra@rowland.harvard.edu}{1}
\addauthor{D. Yamins}{yamins@mit.edu}{2}
\addauthor{D. D. Cox}{davidcox@fas.harvard.edu}{1}

\addinstitution{Rowland Institute at Harvard
    \\100 Edwin H. Land Boulevard
    \\Cambridge, MA 02142, USA}
\addinstitution{Department of Brain and Cognitive Sciences
    \\Massachusetts Institute of Technology
    \\Cambridge, MA 02139, USA}

\runninghead{Bergstra, Yamins, and Cox}{Making a Science of Model Search}

\def\eg{\emph{e.g}\bmvaOneDot}
\def\ie{\emph{i.e}\bmvaOneDot}

\begin{document}

\maketitle

\begin{abstract}

Many computer vision algorithms depend on a variety of parameter choices and settings that are typically hand-tuned in the course of evaluating the algorithm. While such parameter tuning is often presented as being incidental to the algorithm, correctly setting these parameter choices is frequently critical to evaluating a method's full potential. Compounding matters, these parameters often must be re-tuned when the algorithm is applied to a new problem domain, and the tuning process itself often depends on personal experience and intuition in ways that are hard to quantify or describe. Since the performance of a given technique depends on both the fundamental quality of the algorithm and the details of its tuning, it is sometimes difficult to know whether a given technique is genuinely better, or simply better tuned.

In this work, we propose a meta-modeling approach to support automated hyper parameter optimization, with the goal of providing practical tools that replace hand-tuning with a reproducible and unbiased optimization process. Our approach is to expose the underlying expression graph of how a performance metric (\eg classification accuracy on validation examples) is computed from hyper parameters that govern not only how individual processing steps are applied, but even which processing steps are included.  A hyper parameter optimization algorithm transforms this graph into a program for optimizing that performance metric.  Our approach yields state of the art results on three disparate computer vision problems: a face-matching verification task (LFW), a face identification task (PubFig83) and an object recognition task (CIFAR-10), using a single unified algorithm class. More broadly, we argue that the formalization of a meta-model supports more objective, reproducible, and quantitative evaluation of computer vision algorithms, and that it can serve as a valuable tool for guiding algorithm development.

\end{abstract}

\section{Introduction} \label{sec:intro}

Many computer vision algorithms depend on {\em hyper parameter} choices such as the size of filter bank, the strength of classifier regularization, and positions of quantization levels.
These choices can have enormous impact on system performance: \eg
 in
\citep{pinto+cox:2011}, the authors extensively explored a single richly-parameterized model family, yielding classification performance that ranged from chance to state-of-the-art performance, depending solely on hyper parameter choices.
This and other recent work show that the question of ``how good is this model on that dataset?'' is ill-posed.
Rather, it makes sense to speak of the quality of the best configuration that can typically be discovered by a particular search procedure in a given amount of time, for a task at hand.
From this perspective, the tuning of hyper-parameters is an important part of understanding algorithm performance, and should be a formal and quantified part of the scientific method.

On the other hand, \emph{ad hoc} manual tuning by the algorithm inventor, while generally hard to reproduce or compare with fairly, can be efficient.  Since a system's designer has an expectation for how the system should work, he or she can quickly diagnose deviations from that expectation by consulting statistics of that system in operation.

In this work we explore the possibility that manual optimization is no longer efficient enough to justify the lack of formalization that it entails.
Recent developments in algorithm configuration raise the efficiency of automatic search, even in mathematically awkward search spaces, to a level where the result of hand-tuning can be matched and exceeded in a matter of hours on a small cluster of GPU-powered computers.
Using these ideas, we implemented a broad class of feed-forward feature extraction and classification models in order to formalize the steps of selecting the parameters of a model, and evaluating that model on a task.
We compared random search in that model class with a more sophisticated algorithm for hyper parameter optimization, and found that the optimization-based search strategy recovered or improved on the best known configurations for all three image classification tasks in our study.
This success motivates us to suggest that questions regarding the utility of modeling ideas should generally be tested in this style.
Automatic search is reproducible, and thus supports analysis that is impossible for human researchers to perform fairly (\eg ``How would you have tuned approach $Y$ if you had not already learned to optimize approach $X$?'')
To support this kind of research, we provide our automatic hyper parameter optimization algorithm and specification language for download as free open source software.
This software not only completely replicates the research presented in this work, but provides a foundation for general algorithm configuration in future work.

\section{Previous Work} \label{sec:bg}

Our work extends two veins of research with little historical overlap: feed-forward model architectures for computer vision, and techniques for algorithm configuration.

\vspace{12pt}
\noindent \textbf{Feed-forward models in computer vision.}  There is a long tradition of basing computer vision systems on models of biological vision \cite{fukushima:1980,
lecun+etal:1989,
riesenhuber+poggio:1999,
lowe:1999,
hinton+osindero+teh:2006,
dicarlo+zoccolan+rust:2012}.
Such efforts have arrived at a rough consensus model in which nonlinear image features are computed by a feed-forward neural network.
Each layer of the network is comprised of a relatively standard set of transformations, including: (i) dimensionality expansion (\eg by convolution with a filter bank), (ii) dynamic-range reduction (\eg a thresholding), (iii) spatial smoothing (\eg pooling or soft-max), (iv) local competition (\eg divisive normalization), and (v) dimensionality reduction (\eg sub-sampling or PCA).  Feature extraction is usually followed by a simple classifier read-out trained on labeled data.

Beyond this high-level consensus, however, many details remain unresolved: which specific operations should be involved, what order they should be applied in, how many layers should be used, what kinds of classifier(s) should be used, and how (if at all) the filter values should be learned from statistics of input data.
Many competing modeling approaches can roughly be thought of as having made different sets of choices about how to parameterize within a larger unformalized space of feed-forward algorithm configurations.


\vspace{12pt}
\noindent \textbf{Algorithm configuration.}
Algorithm configuration is a branch of optimization dealing with mathematically difficult search spaces, comprising both discrete and continuous variables as well as conditional variables that are only meaningful for some combinations of other variables.
{\em Bayesian} approaches have proven to be useful in these difficult domains.
A Bayesian optimization approach centers on a probability model for $P(\mathrm{score}|\mathrm{configuration})$ that is obtained by updating a prior from a history $H$ of (configuration, score) pairs.
This model can be queried more quickly than the original system in order to find promising candidates.
Search efficiency comes from only evaluating these most promising candidates on the original system.
Gaussian processes \cite{rasmussen+williams:2006} have often been used as the probability model, but other regression models such as decision trees have also proved successful
\cite{MoTiZi78,
hutter:2009,
brochu:2010,
BaKe10,
hutter+hoos+leyton+brown:2011,
bergstra+pinto+cox:2012}.
In these approaches, the criterion of Expected Improvement (EI) beyond a threshold $\mu$ is a popular heuristic for making proposals \cite{Jon01}.
In that approach, the optimization algorithm repeatedly suggests a configuration $c$ that optimizes $EI(c) = \int_{y < \mu} y P(y|c,H)$ while the experimental history of (score, configuration) pairs, $H$, accumulates and changes the model.
Recently \citet{bergstra+bardenet+bengio+kegl:2011} suggested an approach to Bayesian optimization based on a model of $P(c|y)$ instead.
Under some assumptions this approach can also be seen to optimize EI.

Hyper parameter optimization in computer vision is typically carried out by hand, by grid search, or by random search.
We conjecture that Bayesian optimization is not typically used because it is relatively new technology, and because it requires a layer of abstraction between the researcher toying with settings at a command prompt and the system being optimized.
We show that although algorithm configuration is a young discipline, it already provides useful techniques for formalizing the difficult task of simultaneous optimization of many hyper parameters.

\section{Automatic Hyper Parameter Optimization} \label{sec:framework}

Our approach to hyper parameter optimization has four conceptual components:\\

\noindent \textbf{1. Null distribution specification language.}
 We propose an expression language for specifying the hyper parameters of a search space.
 This language describes the distributions that would be used for random, unoptimized search of the configuration space, and encodes the bounds and legal values for any other search procedure.  A null prior distribution for a search problem is an expression $G$ written in this specification language, from which sample configurations can be drawn.

 For example:
 \vspace{-6pt}
 \begin{align} \nonumber G =
 \{a &= \mathbf{normal}(0, 1),\\
\nonumber b &= \mathbf{choice}(0,~\log(\mathbf{uniform}(2,10)),~a) \}
 \end{align}
 specifies a joint distribution in which $a$ is distributed normally with mean 0 and variance 1,
 and $b$ takes either value 0, or $a$, or a value drawn uniformly between 2 and 10.
 There are three hyper parameters at play here, shown in bold: the value of $a$, the value of the choice, and the value of the uniform.

 More generally, the expressions that make up the null distribution specification can be arbitrarily nested, composed into sequences, passed as arguments to deterministic functions, and referenced internally, to form an directed acyclic expression graph (DAG).  \\

\noindent \textbf{2. Loss Function.}
 The loss function is the criterion we desire to minimize.
 It maps legal configurations sampled from $G$ to a real value.
 For example, the loss functions could extract features from a particular image dataset using configuration parameters specified by the random sample from $G$, and then report mis-classification accuracy for those features.  Typically the loss function will be untractable analytically and slow enough to compute that doing so imposes a meaningful cost on the experimenter's time.\\

\noindent \textbf{3. Hyper Parameter Optimization algorithm (HPOA).}  The HPOA is an algorithm which takes as inputs the null prior expression $G$ and an experimental history $H$ of values of the loss function, and returns suggestions for which configuration to try next.  Random sampling from the prior distribution specification $G$ is a perfectly valid HPOA.  More sophisticated HPOAs will generally commandeer the random nodes within the null prior expression graph, replacing them with expressions that use the experimental history in a nontrivial way  (\eg replacing a $\mathbf{uniform}$ node with a Gaussian mixture whose number of components, means, and variances are refined over the course of the experiment). \\

\noindent \textbf{4. Database.}
 Our approach relies on a database to store the experimental history $H$ of configurations that have been tried, and the value of the loss function at each one.  As a search progresses, the database grows, and the HPOA explores different areas of the search space. \\

An important aspect of our approach is the stochastic $\mathbf{choice}$ node, which randomly chooses an argument from a list of possibilities.  Choice nodes make it possible to encode {\em conditional parameters} in a search space.  Visual system models have many configurable components, and entire components can be omitted from a particular pipeline configuration.  The ability to expose to the optimizer how particular configuration variables play no part in a given subtree of choices is crucial to the viability of automatic search in many dimensions.

Our implementation of these four components is available for download as both a general purpose tool for program optimization and a specific optimizable visual system model for new image classification data sets \cite{github-project}.

\section{Object Recognition Model Family}

We evaluate the viability of automatic parameter search by encoding a broad class of feed-forward classification models in terms of the null distribution specification language described in the previous section.
This space is a combination of the work of \citet{coates+ng:2011} and \citet{pinto+doukhan+dicarlo+cox:2009}, and is known to contain parameter settings that achieve the state of the art performance on three data sets (\ie, loss functions):  LFW, Pubfig83, and CIFAR-10.

\begin{figure}
\begin{center}
\includegraphics[width=\textwidth]{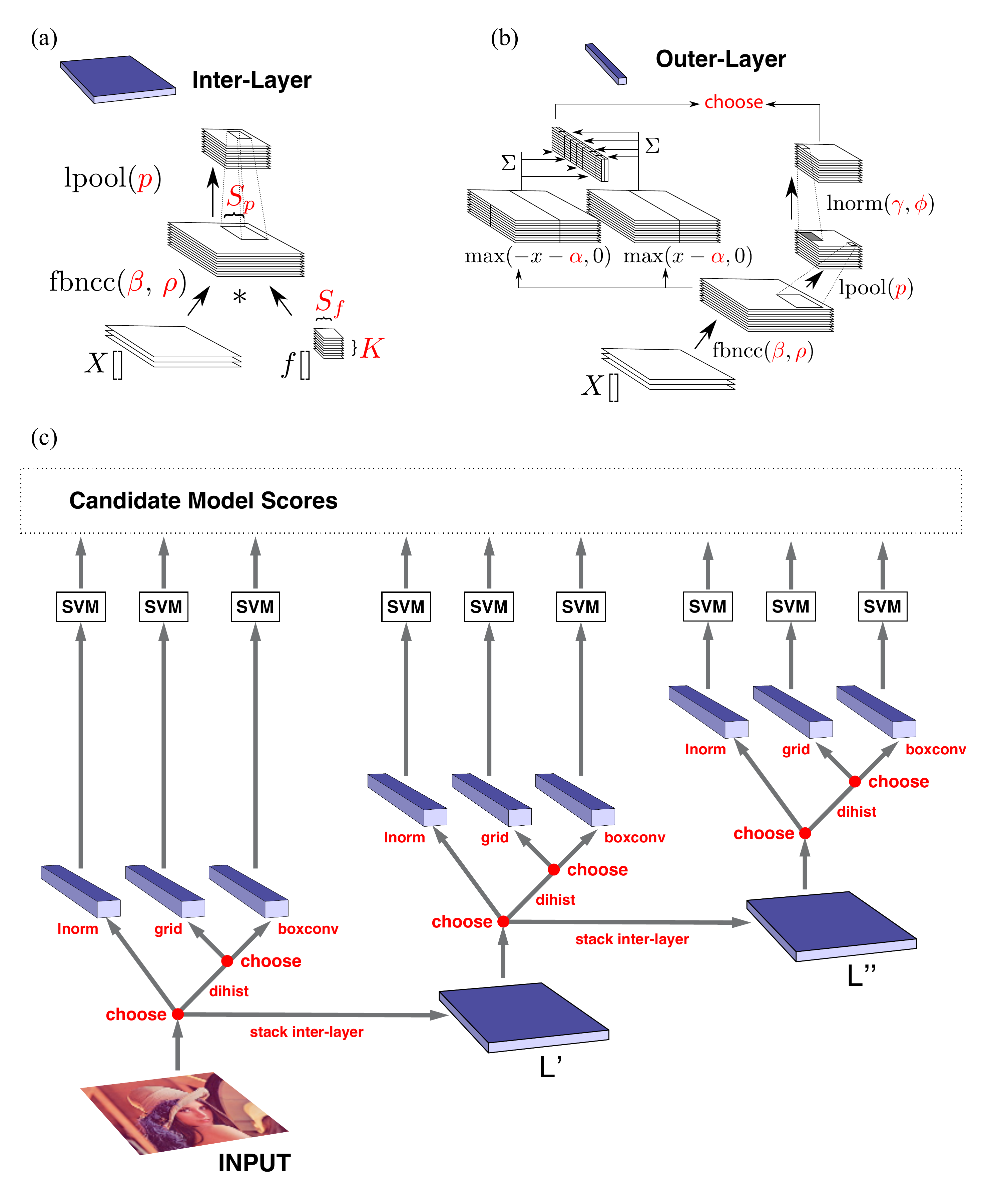}
\caption{
Our experiments search a class of image classification pipelines (c) that include 0, 1, or 2 {\em inter-layers} (a), an {\em outer-layer} (b) that extracts features, and a support vector machine (SVM) classifier.
Hyper parameters govern the number of inter-layers, the type of outer-layer, and a host of configuration options within each processing element.
Although many of the hyper parameters are mutually exclusive (\eg only one outer-layer is active per pipeline) there are over 200 hyper parameters in the full search space.
}
\label{fig:model}
\end{center}
\end{figure}

The full model family that we explore is illustrated in Figure~\ref{fig:model}.
Like \citet{coates+ng:2011}, we include ZCA-based filter-generation algorithms \citep{hyvarinen+oja:2000} and coarse histogram features (described in their work as the R-T and RP-T algorithms).
Like \citet{pinto+doukhan+dicarlo+cox:2009}, we allow for 2-layer and 3-layer sequences of filtering and non-linear spatial pooling.
The remainder of this section describes the components of our model family.
An implementation of the model is available on \citet{github-project}.

The {\em inter-layers} (Figure~\ref{fig:model}a) perform a normalized filter bank cross-correlation, spatial pooling, and possibly sub-sampling.
These layers are very much in the spirit of the elements of the \citet{pinto+cox:2011} model, except that we have combined the normalization and filter bank cross-correlation into a single mathematical operation (fbncc, Equation~\ref{eq:fbncc}).
\begin{align}
y &= \operatorname{fbncc}(x, f) ~~~ \Leftrightarrow ~~~ y_{ijk} = \frac{\check f_k * \check u_{ij}}{\sqrt{\rho \max(||\check u_{ij}||^2, \beta) + (1-\rho )(||\check u_{ij}||^2 + \beta)}} \label{eq:fbncc}
\end{align}
The fbncc operation is a filter bank convolution of each filter $f_k$ with a multi-channel image or feature map $x$, in which each patch $\check x_{ij}$ of $x$ is first shifted by $\epsilon \check m$ (motivating $\check u_{ij} \doteq \check x_{ij} - \epsilon \check m$) then scaled to have approximately unit norm.
Whereas \citet{pinto+cox:2011} employed only random uniform filters $f_k$, we include also some of the filter-generation strategies employed in \citet{coates+ng:2011}: namely random projections of ZCA components, and randomly chosen ZCA-filtered image patches.
Filter-generation is parametrized by a filter count $K \in [16, 256]$), a filter size $S_f \in [2, 10]$, a random seed, and a band-pass parameter in the case of ZCA.
The pair-indexed hat-notation $\check x_{ij}$ refers to a {\em patch volume} from $x$ at row $i$ and column $j$ that includes $S_f$ rows and columns as well as all channels of $x$;
Our fbncc implementation is controlled by
log-normally distributed hyper parameter $\beta$ which defines a low-variance cutoff,
a binary-valued hyper parameter $\rho$ that determines whether that cutoff is soft or hard, and
a binary-valued parameter $\epsilon$ that determines whether the empirically-defined patch mean $\check m$ should be subtracted off or not.

Spatial pooling (lpool, Equation~\ref{eq:lpool}) was implemented as in \citet{pinto+cox:2011}.
\begin{align}
y &= \operatorname{lpool}(x) ~~~ \Leftrightarrow ~~~ y_{ijk} = {x_{i'j'k}} / {||\check x_{i'j'k} ||_p } \label{eq:lpool}
\end{align}
The operation is parameterized by a patch size $S_p \in [2, 8]$, a sub-sampling stride $i'/i = j'/j \in \{1, 2\}$, and a log-normally distributed norm parameter $p$.
The triple-indexed $\check x_{ijk}$ refers to a single-channel {\em patch surface} from $x$ at row $i$, column $j$, and channel $k$ that extends spatially to include $S_p$ rows and columns.

The {\em outer-layers} (Figure~\ref{fig:model}b) combine the fbncc operation of inter-layers with different pooling options.
Rather than sampling or optimizing the filter count, it is determined analytically so that the number of image features approaches but does not exceed sixteen thousand (16,000).
Pooling is done either
(1) with lpool and lnorm (Equation~\ref{eq:lnorm}) as in \cite{pinto+cox:2011}, or
(2) with spatial summation of positive and negative half-rectified filter bank responses (dihist, Equation~\ref{eq:dihist}).
Within pooling strategy (2) we used two strategies to define the spatial patches used in the summation: either (2a) grid cell summation as in \cite{coates+ng:2011}, or (2b) box filtering.
The difference between (2a) and (2b) is a trade-off between spatial resolution and depth of filter bank in making up the output feature set.
\begin{align}
y &= \operatorname{lnorm}(x) ~~~ \Leftrightarrow ~~~ y_{ij} = \begin{cases}\frac{x_{ijk}}{\check x_{ij}} & \text{if} ~ ||\check x_{ij}||_2 > \tau \\ x_{ijk} & \text{otherwise} \end{cases} \label{eq:lnorm} \\
y &= \operatorname{dihist}(x) ~~~ \Leftrightarrow ~~~ y_{ijk} = \begin{bmatrix} ||\operatorname{max}(\check x_{ijk}-\alpha, 0)||_1 \\ ||\operatorname{max}(- \check x_{ijk}-\alpha, 0)||_1 \end{bmatrix}  \label{eq:dihist}
\end{align}
Hyper parameter $\tau$ of the lnorm operation was log-normally distributed, as was the $\alpha$ hyper parameter of dihist.
In approach (2a) we allowed 2x2 or 3x3 grids.
In approach (2b) we allowed for sub-sampling by 1, 2, or 3 and square summation regions of side-length 2 to 8.

The last step in our image-processing pipeline is a {\em classifier}, for which we used an $\ell_2$-regularized, linear, L2-SVM.
For the smaller training sets we used liblinear via sklearn as the solver\cite{fan+chang+hsieh+wang+lin:2008,sklearn}, for larger ones we used a generic L-BFGS algorithm in the primal domain \cite{bergstra+etal:2010}.
Training data were column-normalized.
The classifier components had just two hyper parameters: the strength of regularization and a cutoff for low-variance feature columns.

\section{Results}
\label{sec:results}

We evaluate the technique of automatic hyper parameter configuration by comparing two hyper parameter optimization algorithms: random search versus a Tree of Parzen Estimators (TPE) \cite{bergstra+bardenet+bengio+kegl:2011}.
The TPE algorithm is an HPOA that acts by replacing stochastic nodes in the null description language with ratios of Gaussian Mixture Models (GMM).
On each iteration, for each hyper parameter, TPE fits one GMM $\ell(x)$ to the set of hyper parameter values associated with the smallest (best) loss function values,
and another GMM $g(x)$ to the remaining hyper parameter values.
It chooses the hyper parameter value $x$ that maximizes the ratio ${\ell(x)}/{g(x)}$.
Relative to \citet{bergstra+bardenet+bengio+kegl:2011} we made two minor modifications.
The first modification was to down-weight trials as they age so that old results do not count for as much as more recent ones.
We gave full weight to the most recent 25 trials and applied a linear ramp from 0 to 1.0 to older trials.
The second modification was to vary the fraction of trials used to estimate $\ell(x)$ and $g(x)$ with time.
Out of $T$ observations of any given variable, we used the top-performing ${\sqrt{T}}/{4}$ trials to estimate the density of $\ell$.
We initialized TPE with 50 trials drawn from the null configuration description.
These hyper-hyper parameters were chosen manually by observing the shape of optimization trajectories on LFW view 1 (this is, admittedly, something of a departure from the spirit of this research).


\subsection{TPE vs. Random Search: LFW and PubFig83}

Random search in a large space of biologically-inspired models has been shown to be an effective approach to face verification \citep{pinto+cox:2011} and identification \citep{pinto+stone+zickler+cox:2011}.
Our search space is similar the one used in those works, so LFW \citep{huang+ramesh+berg+learnedmiller:2007} and PubFig83 \citep{pinto+stone+zickler+cox:2011} provide a fair playing fields for comparing TPE with random search.

For experiments on LFW, we follow \citet{pinto+cox:2011} in using the {\em aligned} image set, and resizing the gray scale images to $200 \times 200$.
We followed the official evaluation protocol -- performing model selection on the basis of one thousand images from ``view 1'' and testing by re-training the classifier on 10 ``view 2'' splits of six thousand pairs.
We transformed image features into features of image pairs by applying an element-by-element {\em comparison} function to the left-image and right-image feature vectors.
Following \citet{pinto+cox:2011} we used one comparison function for model selection (square root of absolute difference) and we concatenated four comparison functions for the final ``view 2'' model evaluation (product, absolute difference, squared difference, square root of absolute difference).

The PubFig83 data set contains 8300 color images of size $100 \times 100$, with 100 pictures of each of 83 celebrities \cite{pinto+stone+zickler+cox:2011}.
For our PubFig83 experiments we converted the {\em un-aligned} images to gray scale and screened models on the 83-way identification task using 3 splits of 20 train/20 validation examples per class, running two simultaneous TPE optimization processes for a total of 1200 model evaluations.
Top-scoring configurations on the screening task were then tested in a second phase, consisting of five training splits of 90 train/10 test images per class.
Each of the five second phase training sets of 90 images per class consisted of the 40 images from the first phase and 50 of the 60 remaining images.

\begin{figure}
    \begin{minipage}{\textwidth}
    \begin{center}
        \includegraphics[width=5.0cm]{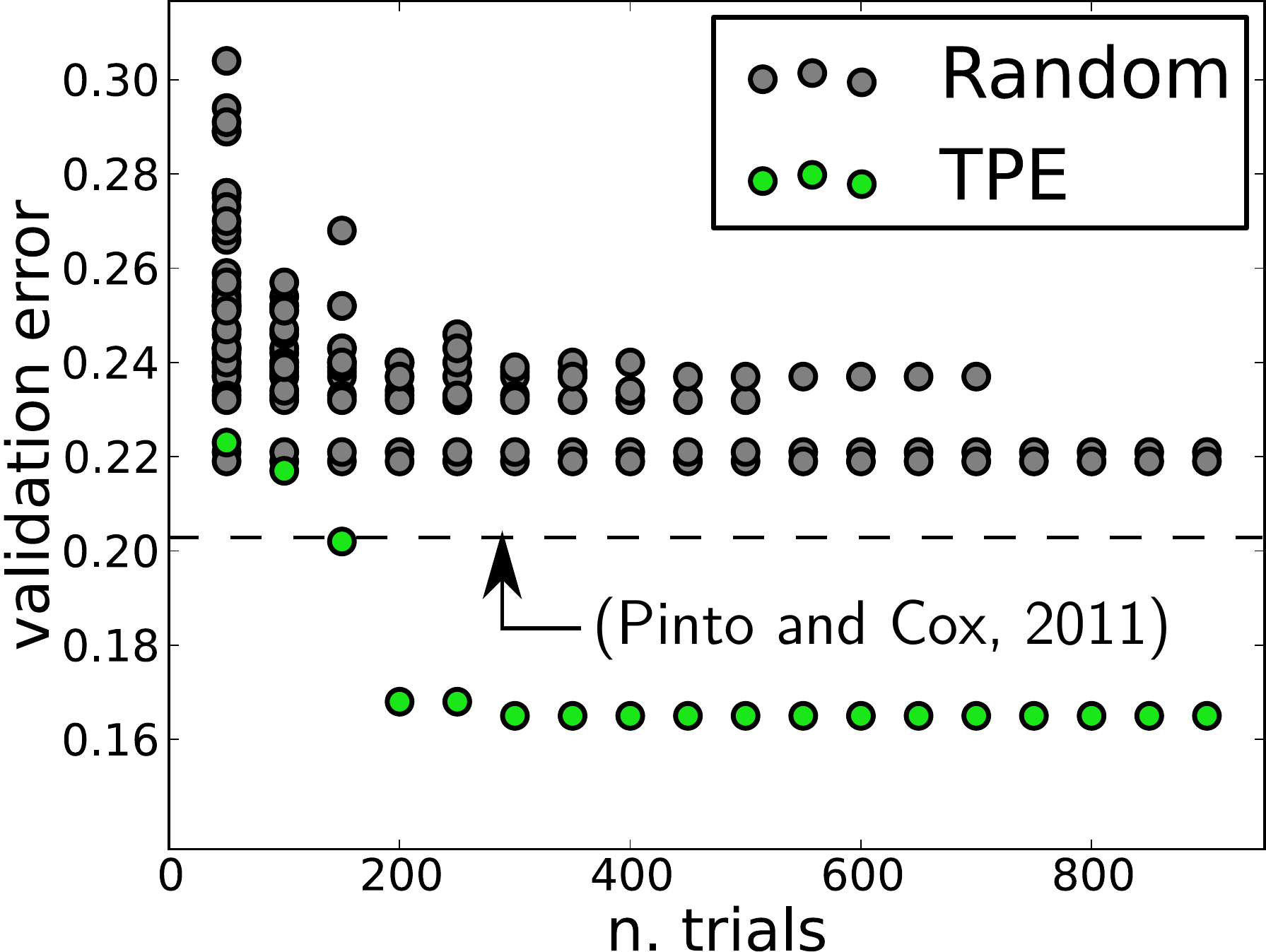}
        \hspace{1cm}
        \includegraphics[width=5.0cm]{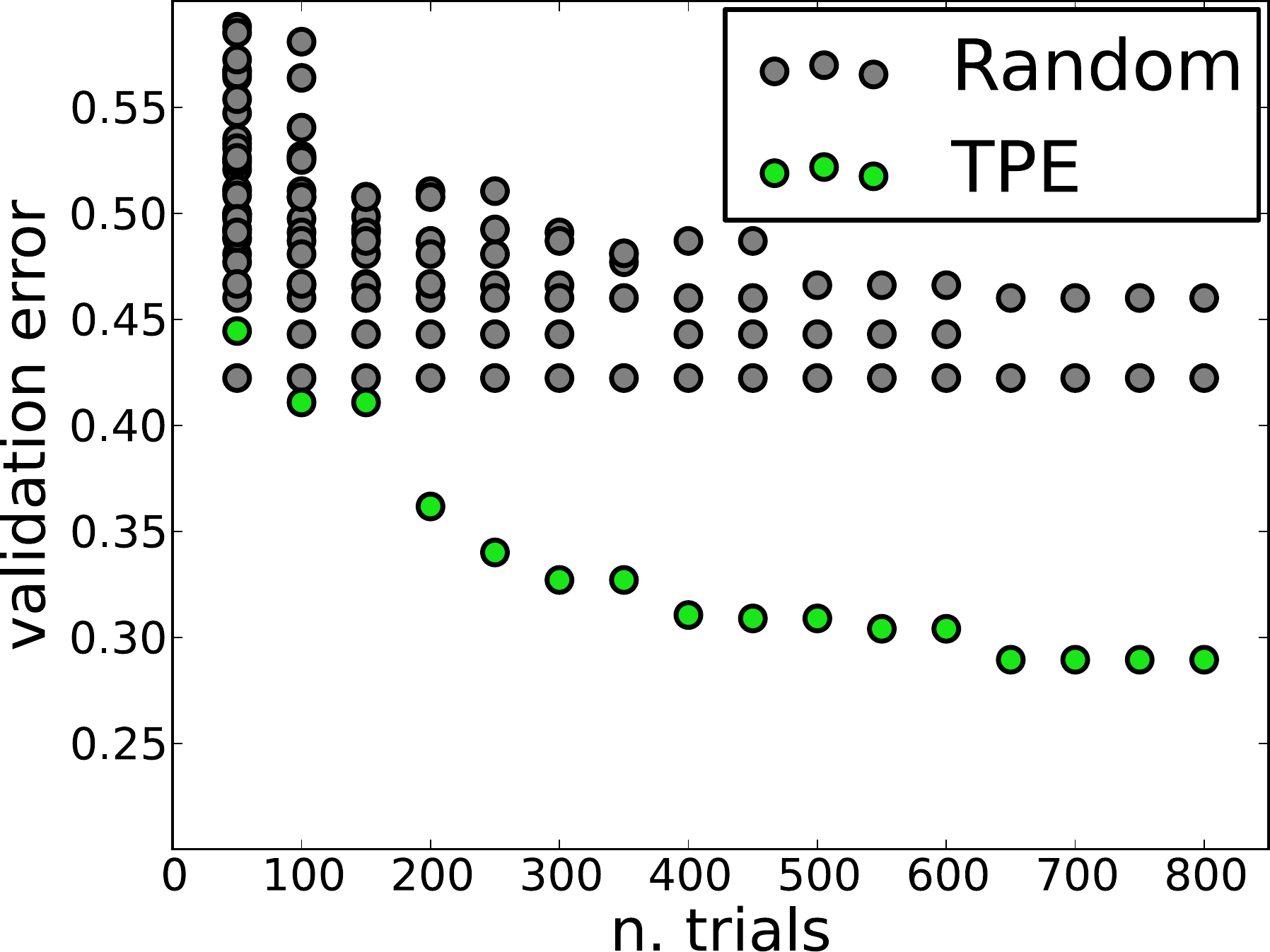}
        \end{center}
    \end{minipage}

    \vspace{.5cm}
    \begin{minipage}{7cm}
        \begin{tabular}{lll}
        \hline
        Method (\# configurations) & LFW View 2 Acc. (\%) & PubFig83 View 2 Acc. (\%) \\
        \hline\hline
        TPE-optimized (750)
            & $\mathbf{84.5} \pm .7$
            & $\mathbf{86.50} \pm .7$
            \\
        High-throughput (15K)
            & $\mathbf{84.1} \pm .7$ \cite{pinto+cox:2011}
            & $85.22 \pm .45$ \cite{pinto+stone+zickler+cox:2011}
            \\
        Random search (2K)
            & $79.2\pm .8$
            & $81.0\pm .8$
            \\
        Chance & 50.0 & 1.2 \\
        \hline
        \end{tabular}
    \end{minipage}
\begin{center}
\caption{
The TPE algorithm finds configurations with significantly better validation set error (top) than a 2000-trial random search or the 15,000-trial random searches carried out in \citet{pinto+cox:2011} and \citet{pinto+stone+zickler+cox:2011}.
Grey dots in the top panels within a column represent the lowest error among $T$ random trials (as $T$ increases to the right); green dots denote the lowest error observed within the first $T$ suggestions by the TPE algorithm.
On test data (``view 2'', bottom), TPE has discovered the best known model configuration in the search space within 750 trials, but our 2000-trial random search has not come close.
View 2 accuracies are given with a 95\% confidence interval assuming Bernoulli-distributed errors.
}
\label{fig:lfw_results}
\label{fig:pubfig83_results}
\end{center}
\vspace{-24pt}
\end{figure}

The results of our model search on LFW are shown in Figure \ref{fig:lfw_results}.
The TPE algorithm exceeded the best random search view 1 performance within 200 trials, for both our random search and that carried out in \citet{pinto+cox:2011}.
TPE converged within 1000 trials to an error rate of 16.2\%, significantly lower than the best configuration found by random search (21.9\%).
On LFW's test data (view 2) the optimal TPE configuration also beats those found by our random search (84.5\% vs. 79.2\%).
The best configuration found by random search in \citet{pinto+cox:2011} does well on View 2 relative to View 1 (84.1\% vs. approximately 79.5\%) and is approximately as accurate as TPE's best configuration on the test set.
On PubFig83, the optimal TPE configuration outperforms the best random configuration found by our random search (86.5\% vs 81.0\%) and the previous state of the art result (85.2\%) \citet{pinto+stone+zickler+cox:2011}.

\subsection{Matching Hand-Tuning: CIFAR-10}

\citet{coates+ng:2011} showed that single-layer approaches are competitive with the best multi-layer alternatives for 10-way object classification using the CIFAR-10 data set \citep{krizhevsky:2009}.
The success of their single-layer approaches depends critically on correct settings for several hyper parameters governing details of the signal processing and feature extraction.
CIFAR-10 images are low-resolution color images ($32 \times 32$) but there are fifty thousand labeled images for training and ten thousand for testing.
We performed model selection on the basis of a single random, stratified subset of ten thousand training examples.

\begin{figure}
\begin{minipage}{6cm}
\hspace{.5cm}\includegraphics[width=5.5cm]{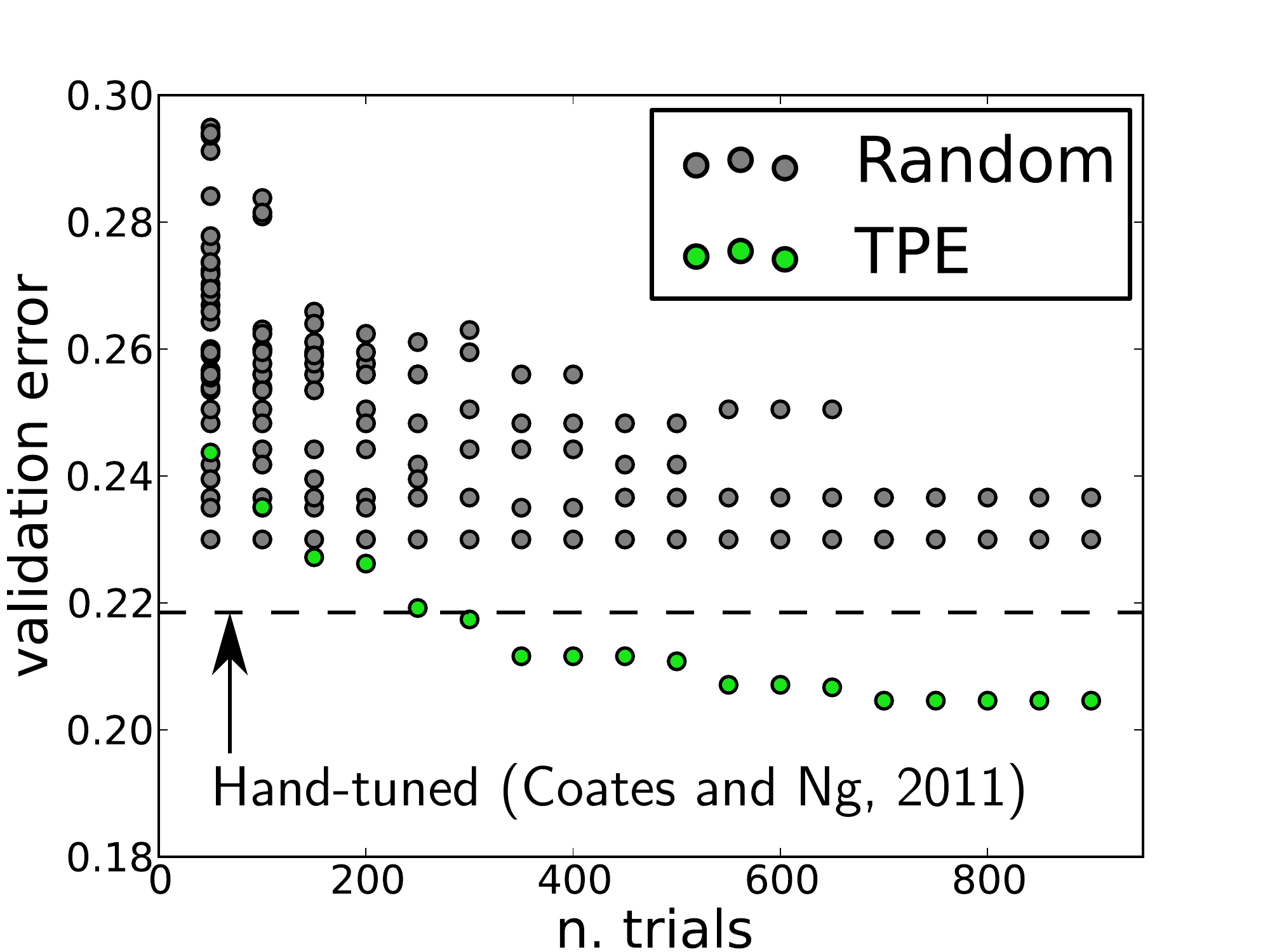}
\end{minipage}
\begin{minipage}{7cm}
    \begin{tabular}{lc}
    \hline
    Method (\# configs) & Test Acc. (\%) \\
    \hline\hline
    Hand-tuned & $\mathbf{79.1}\pm .8$\\
    TPE (800) & $\mathbf{78.8}\pm .8$\\
    Random (2K)  & $76.6\pm .8$\\
    Chance & 10.0 \\
    \hline
    \end{tabular}
    \vspace{.6cm}
\end{minipage}
\begin{center}
\caption{
On the CIFAR-10 object classification data set, TPE minimizes validation set error (left) better than manual tuning and a 2000-point random search.
In test accuracy (right) the best model found by TPE matches the performance of hand-tuning within the model class.
Test accuracies are given with a 95\% confidence interval assuming Bernoulli-distributed errors.
The best configurations from the TPE and random searches are both better on validation than test;
this is normal when the validation set is not perfectly representative of the test set.
}
\label{fig:cifar10_result}
\end{center}
\vspace{-24pt}
\end{figure}

The results of TPE and random search are reported in Figure~\ref{fig:cifar10_result}.
TPE, starting from broad priors over a wide variety of processing pipelines, was able to match the performance of a skilled and motivated domain expert.
With regards to the wall time of the automatic approach,
our implementation of the pipeline was designed for GPU execution and the loss function required from 0 to 30 minutes.
TPE found a configuration very similar to the one found by in \citet{coates+ng:2011} within roughly 24 hours of processing on 6 GPUs.
Random search was not able to match that level of performance, even with as many evaluations.

\section{Discussion} \label{sec:discuss}

In this work, we have described a conceptual framework to support automated hyper parameter optimization, and demonstrated that it can be used to quickly recover state-of-the-art results on several unrelated tasks from a large family of computer vision models, with no manual intervention.
On each of three datasets used in our study we compared random search to a more sophisticated alternative: TPE.
A priori, random search confers some advantages:
    it is trivially parallel,
    it is simpler to implement, and
    the independence of trials supports more interesting analysis \cite{bergstra+bengio:2012}.
However, our experiments found that TPE clearly outstrips random search in terms of optimization efficiency.
TPE found best known configurations for each data set, and did so in only a small fraction of the time we allocated to random search.
TPE, but not random search, was found to match the performance of manual tuning on the CIFAR-10 data set.

This work opens many avenues for future work.
One direction is to enlarge the model class to include a greater variety of components, and configuration strategies for those components.
Many filter-learning and feature extraction technique have been proposed in the literature beyond the core implemented in our experiment code base.
Another direction is to improve the search algorithms.
The TPE algorithm is conspicuously deficient in optimizing each hyper parameter independently of the others.
It is almost certainly the case that the optimal values of some hyper parameters depend on settings of others.
Algorithms such as SMAC \citep{hutter+hoos+leyton+brown:2011} that can represent such interactions might be significantly more effective optimizers than TPE.
It might be possible to extend TPE to employ non-factorial joint densities $P(\text{config} | \text{score})$.
Relatedly, such optimization algorithms might permit the model description language to include distributional parameters that are themselves optimizable quantities (\eg  $\mathbf{uniform}(0, \mathbf{normal}(0,1))$).
Another important direction for research in algorithm configuration is a recognition that not all loss function evaluations are equally expensive in terms of various limited resources, most notably in terms of computation time.
All else being equal, configurations that are cheaper to evaluate should be favored.
Ongoing work such as \cite{snoek+larochelle+adams:2011} is promising but many questions remain.

Our experiments dealt with the optimization of classification accuracy, but our approach extends quite naturally to the optimization (and constrained optimization via barrier techniques) of any real-valued criterion.
We could search instead for the smallest or fastest model that meets a certain level of classification performance,
or the best-performing model that meets the resource constraints imposed by a particular mobile platform.
Having to perform such searches by hand may be daunting,
but when the search space is encoded as a searchable model class, automatic optimization methods can be brought to bear.

\bibliography{paper}

\end{document}